# Effects of Number of Filters of Convolutional Layers on Speech Recognition Model Accuracy

James Mou, Ph.D
*Strategical Solutions, OptumCare*
*UnitedHealth Group*
Eden Prairie, MN, USA
jamesm2199@gmail.com or james.mou@optum.com

Jun Li
*Strategical Solutions, OptumCare*
*UnitedHealth Group*
Eden Prairie, MN, USA
jun.li@optum.com

*Abstract*—Inspired by the progress of the End-to-End approach [1], this paper systematically studies the effects of Number of Filters of convolutional layers on the model prediction accuracy of CNN+RNN (Convolutional Neural Networks adding to Recurrent Neural Networks) for ASR Models (Automatic Speech Recognition). Experimental results show that only when the CNN Number of Filters exceeds a certain threshold value is adding CNN to RNN able to improve the performance of the CNN+RNN speech recognition model, otherwise some parameter ranges of CNN can render it useless to add the CNN to the RNN model. Our results show a strong dependency of word accuracy on the Number of Filters of convolutional layers. Based on the experimental results, the paper suggests a possible hypothesis of Sound-2-Vector Embedding to explain the above observations.

Based on this Embedding hypothesis and the optimization of parameters, the paper develops an End-to-End speech recognition system which has a high word accuracy but also has a light model-weight. The developed LVCSR (Large Vocabulary Continuous Speech Recognition) model has achieved quite a high word accuracy of 90.2% only by its Acoustic Model alone, without any assistance from intermediate phonetic representation and any Language Model. Its acoustic model contains only 4.4 million weight parameters, compared to the 35~68 million acoustic-model weight parameters in DeepSpeech2 [2] (one of the top state-of-the-art LVCSR models) which can achieve a word accuracy of 91.5%. The light-weighted model is good for improving the transcribing computing efficiency and also useful for mobile devices, Driverless Vehicles, and many potential applications requiring light-weighted high-accuracy LVCSR model. Our model weight is reduced to ~10% the size of DeepSpeech2, but our model accuracy remains close to that of DeepSpeech2. If combined with a Language Model, our LVCSR system is able to achieve 91.5% word accuracy.

*Keywords*— *End-to-End Approach, Convolutional Neural Networks (CNN), Number of Filters, Sound-2-Vector Embedding (Convolutional Embedding), Light Model Weight*

## I. INTRODUCTION

"End-to-End" Speech Recognition was published in [1] and some interesting improvements in the End-to-End research area have been made [2, 3, 4, 5, 6]. Traditional ASR (Automatic Speech Recognition) required input feature extractions such as MFCC (Mel Frequency Cepstral Coefficient) [7]. Before the "End-to-End" approach, the deep learning neural networks were only a single component in a complex pipeline of Automatic Speech Recognition models [1]. It was a complex and domain-specific task to build modern large vocabulary continuous speech recognition (LVCSR) systems, which relied on heavily engineered processing stages [3].

### A. Related Works

*1) Study Scope*: This paper systematically studies the effects of Number of Filters of convolutional layers on the model accuracy of CNN+RNN (Convolutional Neural Networks adding to Recurrent Neural Networks) for Speech Recognition.

Scientists [6] have reported that adding CNN to RNN was able to improve the performance of the ASR model, and this study focused on the effectiveness of combining CNN with RNN models. Studies [2, 5] focused on the effects of **Number of Layers** in CNN and RNN, and conv1D on performance. They reported that the WER (Word Error Rate) was improved from 9.52% to 9.20% for 3-layer CNN vs. 1-layer CNN, and was improved from 9.20% to 8.62% for conv2D vs. conv1D. Word accuracy is defined as 100% minus WER. These papers did not study the effects of **Number of Filters** of convolutional layers on model accuracy.

Our work differs from the related works in that we focus on studying the effects of Number of Filters on model accuracy and attempt to find whether there exist any CNN parameters able to act as effective controlling handlers to further optimize the performance of the CNN+RNN architecture model.

Our experimental results show that only when CNN Number of Filters exceeds **a certain threshold value** is adding CNN to RNN able to improve the performance of the CNN+RNN model; but when Number of Filters is below the threshold, adding CNN does not improve model performance. We also find a **strong dependency** of model accuracy on the Number of Filters of convolutional layers, which leads us to consider the possibility of embedding matrix dimension change.

*2) CNN Functionality:* Because of differences in accents, background noises, and speakers, the sound input signals may have wide variations. This fact puts a tough challenge on Speech Recognition. The related works [2, 5, 6] reported adding shallow CNN to RNN Speech Recognition system can improve the word accuracy of the ASR model. However, none of the related works were able to explain the underlying reasons for the observation. While [6] mentioned that CNN may reduce frequency variation for better feature, there was no further discussion or support.

We suggest a possible hypothesis of **"Sound-2-Vector Embedding"**, with CEL (Convolutional Embedding Layer, as described in Chapter II) possibly being able to **effectively extract better feature presentations** from input sound signals with wide variations. These better feature presentations may work more efficiently for the RNN, thus improving the training efficiency and performance of the Speech Recognition systems.

Though we have obtained some experimental results to support this, the hypothesis still requires more experimental results to prove. By sharing this potential hypothesis and its supporting experimental evidences with the academic community of NLP (Natural Language Processing), we hope this can **spark new bright insights** regarding the question of why adding CNN is able to improve CNN+RNN model performance, so that scientists can develop even better architectures for Speech Recognition Models.

*B. LVCSR System Developed*

Based on this hypothesis, the paper develops an End-to-End speech recognition system which has **high word accuracy** but also **light model-weight**. Through optimizing parameters, the developed LVCSR model has achieved quite a high word accuracy of 90.2% by its Acoustic Model alone, without any assistance from any Language Model. In the meantime, we also aim to reduce the model weight (i.e., the number of weight coefficient parameters in the model) to improve the transcribing efficiency once its word accuracy meets a sufficient accuracy standard (e.g., 90%), by parameter optimizations based on our learnings from the experimental results discussed later in this paper. The benefits of a light model-weight for a LVCSR model are significant: it improves the transcribing computing efficiency and can fit into applications with limited computing powers, such as mobile devices and driverless vehicles.

If combined with a Language Model, our LVCSR system can achieve 91.5% word accuracy. In this system, the Sound-2-Vector Embedding is implemented by a CEL (Convolutional Embedding Layer). Although 90% accuracy could be relatively easier for some Image Processing applications or some NLP applications to obtain, it is a tough challenge to develop a large vocabulary continuous speech recognition model which reaches **word accuracy of 90%** on LibriSpeech dataset but also retains its **Model Weight under 10-million weight** coefficient parameters, like the proposed LVCSR model in this paper.

## II. SOUND-2-VECTOR EMBEDDING

The Sound-2-Vector Embedding is a Speech Recognition analogy to the embedding process of **word2vec** [8, 9, 10] in the NLP area. The word2vec model includes the skip-gram (SG) model and the continuous bag-of-word (CBOW) model. It is a shallow neural network which is trained to reconstruct linguistic contexts. It contains an Input Layer, a Hidden Layer, and an Output Layer [9]. The Hidden Layer of word2vec was designed to be a Dense Layer typically containing 300 or more simple neurons [8, 10]. The word2vec model used a **Dense Layer** [9] and never used a Convolutional Layer as its Hidden Layer.

Based on our experimental results, our hypothesis in this paper entails one Convolutional Layer (CNN) working as the Hidden Layer and the weight coefficients of the Filters of the convolutional layer acting as the embedding matrix. Our paper differs from word2vec [8, 10] in that we use a **Convolutional Layer** as its **Hidden Layer** for embedding (instead of a Dense Layer as its Hidden Layer in word2vec)**.**

Therefore, the **"Sound-2-Vector Embedding"** is also referenced as **"Convolutional Embedding".**

Our results show that not all CNNs with any parameters can improve word accuracy. Only the subset of CNNs with appropriately-designed parameters are able to have the embedding effect and improve word accuracy when added to RNN speech recognition model. Thus we define this specific subset of CNNs with appropriate parameters as *Convolutional Embedding Layer* (CEL). For typical ranges of workable parameters for the Convolutional Embedding Layer, please see discussions in Section IV.B.

We hypothesize two possible architectures for the Sound-2-Vector Embedding in the Speech Recognition model as shown in Fig. 1 and Fig. 2, motivated by word2vec [8, 10]. Our hypothesized Embedding architecture consists of one Input Layer, one Hidden Layer (using Convolutional Layer), and one Output Layer.

*A. Character Set Embedding (CSE) Model*

The embedding architecture is shown in Fig. 1.

- The Input Layer is the Sound Input Preprocessing Layer which conducts Fourier Transformations and sliding window operations, resulting in *Spectrograms* of the sound signals. Assuming we pick up a fixed number of context frames in the Study Window (defined as a few neighbor *context frames before and after* the frame in time series), the dimension of the Study Window of the Input Layer is denoted as S. Input $x=\{x_1, x_2, …, x_S\}$.

- The Hidden Layer is a 1-layer CNN and its convolutional layer is expected to fully utilize the trainable weight coefficients in its sufficient Number of Filters to store the information of the Embedding Matrix. Its dimension is denoted as N.

- The Output Layer contains the Output Target Values of the Multiclass Classifications, which reflect whether the input sound signal represents one of the characters inside the Model Final Target Set. The output is a one-hot encoded vector with M dimension. $y=\{y_1, y_2, …, y_M\}$.

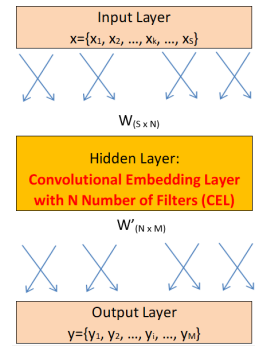

Fig. 1. Illustration of the architecture of the Character Set Embedding (CSE) Model.

Because we use the CTC (Connectionist Temporal Classification) Approach, we only target the characters in English. For example, we use 30 Single Characters in our Model Final Target Set (including 26 English letters, plus blank, space, <UNK> and apostrophe). The dimension M of the Output Layer is defined by the size of the Model Final Target Set.

The weights between the Input Layer and the Hidden Layer can be represented by a (S × N) matrix W [9], and the weights between the Hidden Layer and the Output Layer can be represented by a **different** matrix W**'** with dimensions of (N × M). To comply with the convention of Embedding, we define the Embedding Matrix E as the Transpose of the above (S × N) Weight Matrix W between the Input Layer and the Hidden Layer, so E is a matrix with dimensions of (N × S):

$$E = W^T \quad (1)$$

We can set Convolution Filter Size = S (the dimension of the Study Window of the Input Layer), so that the convolution operation results in a 1-dimension vector. Then the Hidden Layer Vector h for the Study Window is defined as:

$$h_{(N\times 1)} = H_{convol}(E_{(N\times S)}, x_{(S\times 1)}) \quad (2)$$

where $H_{convol}()$ is the function of the Convolution Operation applied to E and x, E is the Embedding Matrix and x is the vector of inputs, and h is the vector of the resulting values (convolved features) in the Hidden Layer by the convolution of the input vector x with the Embedding Matrix E. The equations and derivations **only aim to elucidate the concept** that the "size of the Embedding Matrix" of the Sound-2-Vector Embedding is proportional to the "Number of Filters" of the convolutional layer.

Similar to the skip-gram model [8] in the word2vec, the activation function between the Input Layer and the Hidden Layer is designed to be linear, but the activation function between the Hidden Layer and the Output Layer is designed to be nonlinear. The latter activation function can be 'softmax', 'elu' [11], or 'relu'. For example, we use softmax as the activation function with its own different weight coefficients W' of a (N × M) matrix. Thus, if we ignore the bias term, we can get the probability distribution [9] of a single character $y_i$

$$P(y = y_i | x) = \frac{exp((W'_i)^T * h)}{\sum_{j=1}^{M} exp((W'_j)^T * h)} \quad (3)$$

where $W'_i$ and $W'_j$ are the i[th] and the j[th] column vectors of the matrix W' respectively, W' is the weight matrix between the Hidden Layer and the Output Layer, h is defined in (2) and T = transpose operation. The Loss Function can be written as

$$L(y, y\_hat) = -\sum_{i=1}^{M}(y_i \times \log(y_i\_hat)) \quad (4)$$

where $y_i\_hat$ is the model predicted value and $y_i$ is the ground truth value in the labels (the character set) respectively.

After the above embedding model has been well trained, the Embedding Matrix E will then be able to reflect the mapping between the input values to the Model Final Target Set in the Output Layer. For the Convolutional Embedding Layer with sufficient Number of Filters as the Hidden Layer, the Embedding Matrix E is stored as the matrix of weight coefficient parameters in the Filters of the CEL.

B. *Convolutional Recurrent Neural Network Embedding (CRNNE) Model*

The embedding architecture of CRNNE is shown in Fig. 2. The architecture of CRNNE is similar to that of the CSE Model, except that **its Output Layer is a *virtual* layer**. The CSE Model belongs to the Pre-Defined Embedding style, which is the need to pre-define the Target Set in the Output Layer and use the above Embedding Structures to pre-train the Embedding Matrix outside of the system of interest ahead of time, and then insert the resulting Embedding Matrix into the system for Transfer Learning purposes like the word2vec.

We can also consider another option of Self-Adaptive Embedding style, which is letting "Machine Learning" learn the appropriate Target Set for the Output Layer, instead of manually predefining {$y_1$, $y_2$, …, $y_k$, …, $y_M$} before the embedding training. This is another type of "End-to-End" approach, because we only care about the Input End and the Output End. As long as the middle hidden weight coefficients can do a good mapping of the Input to the Output, we do not mind whether the explicit content of the appropriate target set in the Output Layer is interpretable by humans, and we **let Machine Learning learn** a *better form of Feature Representations* as the Output Layer content, which is then fed to the RNN for sequence modeling.

The Convolutional Recurrent Neural Network Embedding (CRNNE) Model is a good example of this idea. The dimension of the Virtual Output Layer is the same as the number of neurons in the first layer of RNN. We insert the CRNNE Embedding Structure into the system of interest and then allow the CNN (CEL) and RNN to train collaboratively via the forward propagations and the backward propagations for many epochs. The experimental results are quite promising with this approach.

The advantage of the Self-Adaptive Embedding style is that humans do not have to **manually predefine** what should be {$y_1$, $y_2$, …, $y_k$, …, $y_M$}, and Machine Learning will retrieve better forms of Feature Representations as the Output Layer content for the RNN. By allowing the CEL and the RNN (its backward propagation) to closely communicate and negotiate through many epochs of collaborations, the final common Communication Protocol Interface that has been negotiated by the CEL and the RNN can be quite efficient for a powerful Embedding Effect. Please note the Common Communication Protocol format will be meta-target and may be **uninterpretable** for humans at all, since it is of some form of machine encoding.

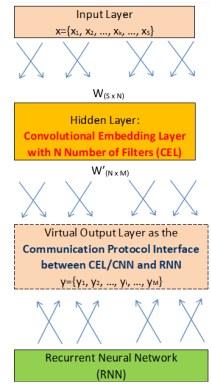

Fig. 2. Illustration of architecture of the Convolutional Recurrent Neural Network Embedding (CRNNE) Model.

This type of synergy between CEL (CNN) and RNN allows the whole CNN+RNN Speech Recognition Model to achieve better word accuracy than the RNN-only model without CEL.

III. EXPERIMENTS

*A. Model Architecture and Details*

The dataset used in this study is the LibriSpeech Clean Set (courtesy to Dr. Daniel Povey at Johns Hopkins University). We manually review the whole dataset and exclude those samples with the issue of mismatching transcription with audio file. To assure the same distribution, we mix the samples and then randomly split the dataset into Train/Dev/Test Dataset at the ratio of 8:1:1. The Dev Dataset is used during model selection/tuning. The WER (Word Error Rate) and CFV (Cost Function Value) data results for using the Dev Dataset are almost the same as the WER and CFV data for the Test Dataset in this study since they have the same distribution by design. We here only report the WER and CFV data results of using the Test Dataset in this paper. That is to say, the term Validation Dataset in this paper refers to the Test Dataset.

To make later discussions easier, this developed LVCSR Model is referred to as the "**CRNN Model**" in this paper, because we try to combine the benefits of CNN and RNN. The architecture of the CRNN Model is as simple as in Table I.

This implementation reflects the Self-Adaptive Embedding style, i.e., the CRNNE (Convolutional Recurrent Neural Network Embedding) as discussed in Section II.B. Our work differs from Google's CLDNN [6] in that our model does not have a DNN (Deep Neural Network) after RNN, i.e., we only combine CNN+RNN. We applied the Nesterov SGD Algorithm for all Model Trainings.

*B. Preliminary Result for Model Evaluation*

Fig. 3 shows a preliminary result that can help us quickly estimate the final WER performance of different models with various parameter combinations for pre-screening in the study.

When models use the End-to-End approach and the CTC as its cost function, we observe that the Final Word Error Rate demonstrates a strong positive correlation with its corresponding CTC Cost Function Value of the same Validation Dataset for acoustic models with the same CNN+RNN architecture but with various CEL hyperparameters.

This agrees with the initial intent of the End-to-End design [1]. The rule in Fig. 3 will help us quickly estimate the potential final WER performance of different models by simply comparing their corresponding Cost Function Value changes within the first 20 training epochs, without having to train each of the different models to completion and then measuring WER, in order to evaluate how different hyperparameters affect model accuracy. The latter can be very time-consuming because complete training of one speech recognition model takes a long time, and we have so many possible combinations of hyperparamters to study in our DOEs (Design of Experiment) or Grid Searches. Using the changes of Cost Function Value (CFV) to assist the pre-screening of parameters for DOEs saves us a significant amount of time in parameter pre-screening. After selections, we have completely trained all the selected models and measured their WER (for word accuracy) as our **Final Evaluation** of the performance of the models in this study.

IV. RESULTS, ANALYSES AND DISCUSSIONS

Sections A and B are discussions about the Effects of Number of Filters of convolutional layer on model performance, Sections C, D and E present the word accuracy and model weight of the CRNN model developed in this paper, and Section F summarizes the comparison of the related works.

*A. Effect of Convolutional Embedding Layer (CEL) Parameters on Model Training Dynamics*

Because of the restriction of paper space, we only show a subset of our extensive experiments. Fig. 4 demonstrates the Training Dynamics of the CTC Cost Function Value (CFV) of models using different combinations of Kernel Size (some papers call Filter Size [4]) and Number of Filters of the CEL.

TABLE I. THE ARCHITECTURE OF THE CRNN MODEL

| Layers of the CRNN Model | Parameters |
|---|---|
| Sound Input Preprocessing Layer | To 81 Frequency Bins as Input |
| Convolutional Embedding Layer (CEL) | Number of Filter = 200, Kernel size = 11, stride = 2, padding = 'valid' |
| Bidirectional GRU | Dropout = 0.25 |
| Bidirectional GRU | Dropout = 0.25 |
| Bidirectional GRU | Dropout = 0.25 |
| Bidirectional GRU | Dropout = 0.25 |
| Softmax Layer | Target at the 30-character set |

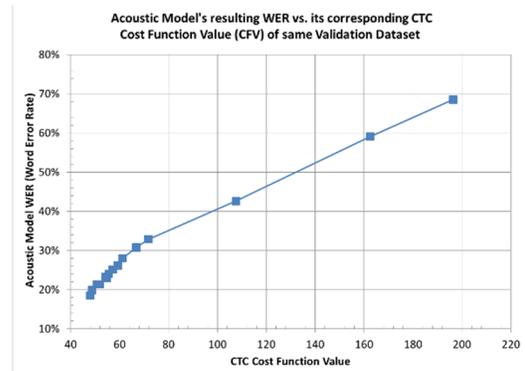

Fig. 3. The Word Error Rate (WER) vs. its corresponding CTC Cost Function Value of the same validation dataset for acoustic models with same CNN+RNN architecture but with various hyperparameters of CEL (Dropout = 0).

When using the same Number of Filters of CEL, two different Kernel Size curves almost always overlap, which indicates that the Kernel Size of the Convolutional Embedding Layer makes almost no difference in the Training Dynamics of the models.

But with the same Kernel Size of CEL, two different Number of Filters curves have a wide gap in CFV, which indicates that **Number of Filters** of the Convolutional Embedding Layer *makes a large difference* in the speed of the Cost Function Value reduction (i.e. Training Dynamics).

The above observations make sense because the Number of Filters of the convolutional layer actually is proportional to the size of the Embedding Matrix of the Sound-2-Vector Embedding. The increasing Number of Filters of the CEL will bring in more numbers of trainable Weight Coefficients to the Embedding Matrix, thus the Embedding Matrix can better extract more plentiful features from the sound inputs. Therefore, the increasing Number of Filters results in better training efficiency and higher word accuracy of the model. In order to confirm this hypothesis, we measured more different models with various Number of Filters of the CEL ( as shown in Fig. 5).

Fig. 5 shows the CTC Cost Function Values just after the 20th epoch training (snapshot) for different models. The results confirm that the models with the larger Number of Filters of the Convolutional Embedding Layer do have much more efficient sound-2-vector embedding effects and result in lower Cost Function Values and lower Word Error Rate.

When the Number of Filters of Convolutional Embedding Layer is less than 10, because the number of rows in the embedding matrix is not sufficient to embed enough useful sound features, its sound-2-vector embedding effects are too weak and are negligible. The CTC Cost Function Values are still

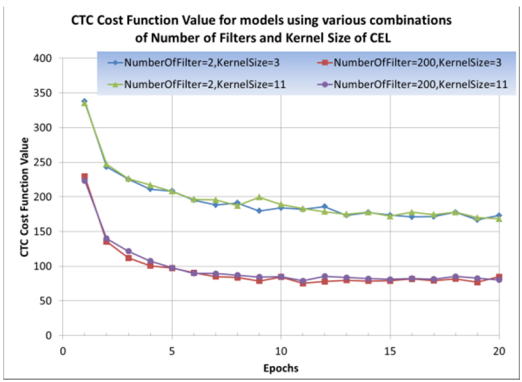

Fig. 4. The CTC Cost Function Value (CFV) changes during the model training process vs. the Kernel Size and the Number of Filters of the Convolutional Embedding Layer used in the models, with Padding Border Mode = "valid".

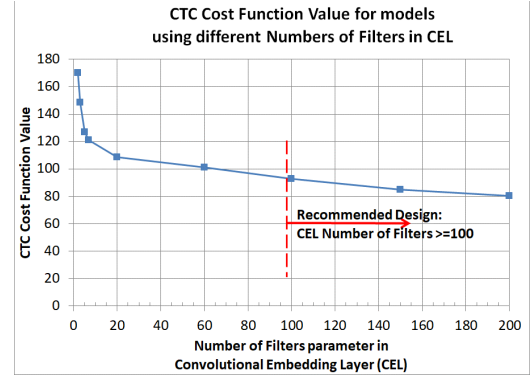

Fig. 5. The CTC Cost Function Value (CFV) for models vs. the Number of Filters of the Convolutional Embedding Layer used in the models, each data point is taken after 20 epochs of training, with Kernel Size = 11 and Padding Border Mode = "valid".

higher than 120 for this group which consequently does not have good accuracy as expected.

When the Number of Filters of Convolutional Embedding Layer is between 20 and 200, the CTC Cost Function Value after 20 epochs training continuously decreases with increasing Number of Filters.

The above observations suggest that there does exist a Threshold Number of Filters value (i.e., the Embedding Matrix has a Minimum Size value) in order for the sound-2-vector embedding to take effect. The Threshold Number of Filters value is about 10 for the sound-2-vector embedding system in this experiment. If you desire a more effective sound-2-vector embedding in your model, then a Number of Filters of 100 or more is recommended for the Convolutional Embedding Layer (also see Fig. 7). The larger the Number of Filters of the CEL is, the better performance the model will achieve.

In addition, experimental results also confirm that the Training Dynamics (CTC CFV behaviors) for the Padding Border Mode of "valid" vs. "same" are almost identical to each other. The reason for the observations is that the Padding Border Mode itself cannot change the size of the Embedding Matrix, which is similar to the above Kernel Size Effect.

### B. Discussion of the Effects of Number of Filters of Convolutional Layers on Model Accuracy

The (1) ~ (4) that were derived in Section II.A. are generalized and can be applicable to both the CSE Model and the CRNNE Embedding Models proposed in this article.

In (1), the Embedding Matrix E is defined as the transpose of the (S × N) Weight Matrix W between the Input Layer and the Hidden Layer. The E has the following structure:

$$E = \begin{bmatrix} e_{11} & \cdots & e_{1S} \\ \cdots & \cdots & \cdots \\ e_{N1} & \cdots & e_{NS} \end{bmatrix}_{(N \times S)} \quad (5)$$

The Embedding Matrix E has N rows and S columns, where S is the dimension of the Study Window of the Input Layer and N should be proportional to the Number of Filters of the convolutional layer. The number of rows N in the matrix is defined as the **Depth of the Embedding Matrix**, as this Depth parameter is controlling the size of the Embedding Matrix E, which decides how many feature maps can be captured from the Sound Inputs. This key parameter can change the training behaviors of the model with such an Embedding CEL structure.

Like word2vec, a word embedding with a small dimension is typically not expressive enough to capture all possible word relations [12, 13]. The same rule should also be applied to the sound-2-vector embedding matrix, because more rows in the Embedding Matrix can allow it to capture increasing numbers of plentiful feature representations from the sound inputs. The experimental results confirm that the larger the Number of Filters of the CEL, the better the word accuracy that the model will obtain.

This Embedding Hypothesis is able to explain the following behaviors in our experiments:

- The Number of Filters of Convolutional Embedding Layer control the Training Dynamics of model, such as how faster the Cost Function Value (CFV) will be reduced with epochs. The model with larger Number of Filters reduces the CFV faster than the similar model with smaller Number of Filters, and results in better word accuracy because of its better feature extractions of CEL.

- There exists a **threshold value for the Number of Filters** parameter. When below the threshold value, the Sound-2-Vector Embedding is too weak and has an almost negligible embedding effect; when above the threshold value, the Sound-2-Vector Embedding effect is significant and the Embedding Efficiency increases with the increasing Number of Filters. In other words, Embedding Matrix has a Minimum Depth Value in order for the sound-2-vector embedding to take effect.

### C. The Word Accuracy of the Model with Convolutional Embedding Layer vs. the Model without Convolutional Embedding Layer

Fig. 6 compares the training efficiency of the Speech Recognition Model with the Convolutional Embedding Layer (CEL) vs. the Model without Convolutional Embedding Layer.

The CTC Cost Function Value of the Model with a CEL structure is reduced significantly faster than that of the Model without CEL. The final word accuracy of the acoustic model with CEL also has a significant improvement over the acoustic model without CEL, if its Number of Filters of convolutional layers is larger than or equal to 100, as shown in Fig. 7.

CEL (convolutional layer) utilizes its filters' weight coefficients (i.e., the Embedding Matrix) to extract more useful feature representations from raw sound signals and feed the better extracted-features as the inputs directly to the RNN. The experimental results confirm that for ASR models with the "CEL + RNN" architecture where CEL = (shallow CNN), the sound-2-vector embedding effect of the CEL may possibly contribute to improve the model performance, as discussed in Section IV.B. This paper only focuses on "(shallow CNN) + RNN" architecture ASR models. (For ASR models with "Deep CNN Only" architectures [4, 14] which have not been discussed in this paper, there may be some other effects besides the sound-2-vector embedding effect to enhance performance.)

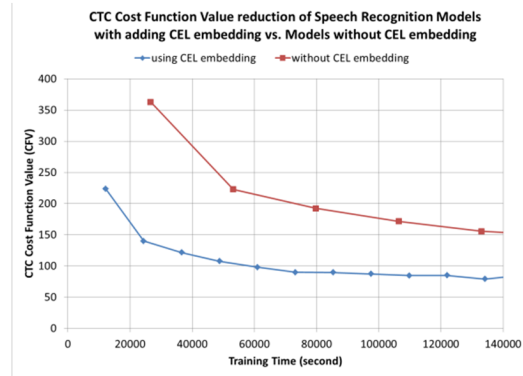

Fig. 6. CTC Cost Function Value (CFV) changes of the model with Convolutional Embedding Layer vs. the model without Convolutional Embedding Layer.

### D. The Word Accuracy Performance of the Final Selected Acoustic Model

Fig. 8 shows that the final Word Accuracy of the Validation Dataset for the Acoustic Model Only with Dropout vs. without Dropout. If we only look at the Acoustic Model (without any lifting from Language Model), the final Word Accuracy of the "CEL+RNN" Acoustic Model with Dropout of 0.25 can achieve 90.2%. However, the "CEL+RNN" Acoustic Model without Dropout can only reach 81.5%. The relative improvement for the Model with Dropout of 0.25 is = (90.2% - 81.5%) / (100% - 81.5%) = 47%, which is very significant.

Now let us shift attention to the performance of the CRNN Model (i.e., the "CEL+RNN" Acoustic Model with Dropout of 0.25) combined with Language Model: its final Word Accuracy is pushed up to 91.5%.

We give the following example from the Test Dataset (i.e., the Validation Dataset) to demonstrate the performance of our CRNN Acoustic Model alone (without any lifting from dictionary or language model). The four error words (highlighted in **Bold**) out of total 52 words give a Word Error Rate (WER) of 7.8% and a Word Accuracy of 92.2%.

- The Ground Truth Transcription Texts are:

a blast of fire sprayed the ground **then** turned off we have **four** minutes to the next one we hit the long period they ran stumbling in the soft ashes tripping over charred bones and rusted metal two **men grabbed** jason under the arm and half carried him across the ground

- The Transcription Texts that are predicted by our CRNN Acoustic Model alone are:

a blast of fire sprayed the ground **than** turned off we have **for** minutes to the next one we hit the long period they ran stumbling in the soft ashes tripping over charred bones and rusted metal two **man grab** jason under the arm and half carried him across the ground

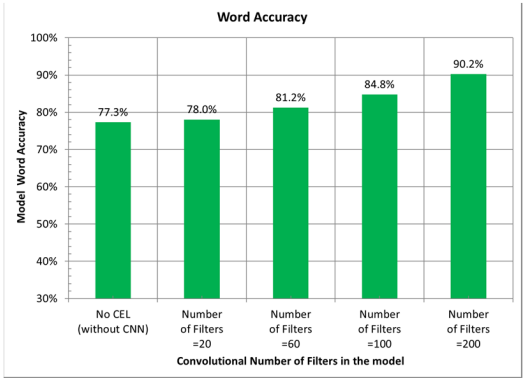

Fig. 7. Word Accuracy of the model without Convolutional Embedding Layer vs. the model with Convolutional Embedding Layer, and the effects of **Number of Filters**. Each data point is the final word accuracy for Acoustic Model only (without Language Model) with Dropout = 0.25 for different models.

### E. Light Model Weight Discussion

One additional advantage of the above CRNN Model with Dropout = 0.25 is that its acoustic model contains only **4.4 million weight parameters** (compared to 35~68 million acoustic-model weight parameters in [2] and 23 million weight parameters in [14]). Therefore, this Sound-2-Vector Embedding End-to-End Speech Recognition Model is a light-weighted ASR Model (as shown in Fig. 9), which is good for the Real Time Factor and allows for it to be easily loaded into mobile devices, "driverless vehicles", and many potential applications requiring **light-weighted high-accuracy** LVCSR models.

With our model weight being reduced to ~10% size of the DeepSpeech2, the transcribing speed is almost proportionally increased to be 7 times faster, or the same server capacity can process 7 times more transcription requests concurrently. The word accuracy is at almost the same level: 90.2% vs. 91.5%.

### F. Comparison of the Related Works

Table II summarizes the comparison of the Word Accuracy performance and the Acoustic Model Weight (i.e., the Number of Weight Coefficient Parameters in the acoustic model) of this CRNN model vs. the related works (DeepSpeech2 [2, 5] from Baidu Research and CLDNN [6] from Google). We bring in the CNN-based Features model [15] from Google as a baseline since it attempted to combine CNN only with DNN (Deep Neural Networks) but without any RNN.

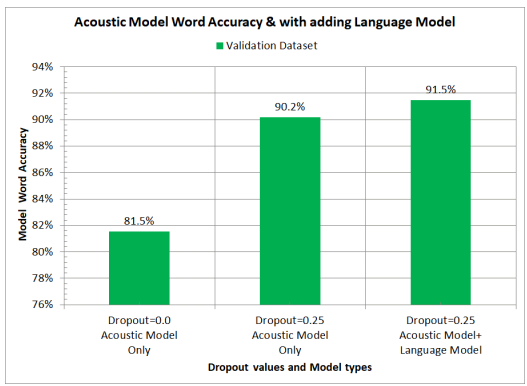

Fig. 8. Final Word Accuracy of the Acoustic Model with Dropout = 0.25 vs. the Acoustic Model with Dropout = 0.

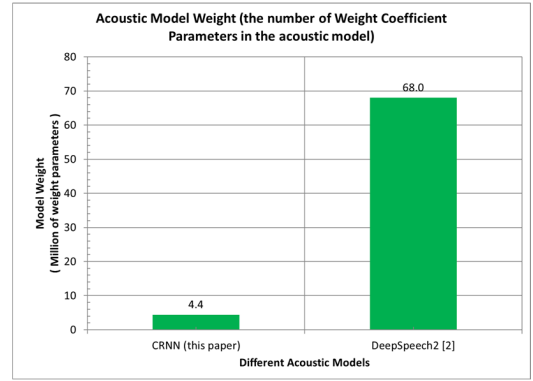

Fig. 9. Comparison of the Acoustic Model Weight (the number of Weight Coefficient Parameters in the acoustic model). DeepSpeech2 model weight data are retrieved from the Table 3 in DeepSpeech2 [2].

Table III compares different study scopes and different optimization parameter which this paper is concentrated in vs. the related works.

CLDNN [6] and DeepSpeech2 [2] have set good examples for implementing End-to-End speech recognition models. This paper was inspired by these related works. Our work differs from theirs in that we systematically study the effects of Number of Filters of convolutional layers on speech recognition model accuracy. We find that only when CNN Number of Filters exceeds a certain threshold value is adding CNN to RNN able to improve the performance of the CNN+RNN speech recognition model. We also find a strong dependency of model accuracy on the Number of Filters of convolutional layers.

TABLE II. **WORD ACCURACY** PERFORMANCE AND THE **ACOUSTIC MODEL WEIGHT** (I.E., THE NUMBER OF WEIGHT COEFFICIENT PARAMETERS) OF THIS CRNN MODEL VS. THE RELATED WORKS

| Is Related Works? | Model and paper reference | WER of Acoustic Model | Acoustic Model Word Accuracy | Model Weight (number of Parameters) |
|---|---|---|---|---|
| Related Works | CNN-based Features [15] | 15.0% [a] | 85.0% | N/A |
| Related Works | CLDNN [6] | 13.1% [b] | 86.9% | N/A |
| Related Works | DeepSpeech2 [2, 5] | 8.5% [c] | 91.5% | **68 million** [c] |
| This paper | CRNN | 9.8% | 90.2% | **4.4 million** |

[a]. Data are retrieved from the Table 5 in CNN-based Features [15]

[b]. Data are retrieved from the Table 8 in CLDNN [6]

[c]. Data are retrieved from the Table 3 in DeepSpeech2 [2]

TABLE III. COMPARISON OF DIFFERENT STUDY SCOPES AND OPTIMIZATION PARAMETER OF THE RELATED WORKS

| Model Name | Main Study Scope | Optimization Parameters | Did it find that "**some parameter ranges** of CNN can render it **useless** for adding CNN to RNN speech recognition model" ? |
|---|---|---|---|
| CNN-based Features [15] | combine CNN + DNN | combined structure | N/A |
| CLDNN [6] | combine CNN + LSTM + DNN | combined structure | **no** |
| DeepSpeech2 [2, 5] | **Number of Layers** of CNN & RNN effect on accuracy | Number of **Layers** of CNN, etc. | **no** |
| CRNN | **Number of Filters** of CNN effect on accuracy | Number of **Filters** of CNN, Dropout, etc. | **Yes,** we find the non-working parameter range and the threshold value |

## V. CONCLUSIONS

Experimental results confirm that the **Number of Filters** of CEL (Convolutional Embedding Layer) is a key parameter that is able to significantly change the training dynamics and the word accuracy of the CEL+RNN models.

The Number of Filters of CEL is proportional to the Size of the Embedding Matrix, which controls the efficiency of Feature Extractions of the Sound-2-Vector Embedding.

There exists a **Threshold Number of Filters of CNN** below which can render it useless to add CNN to RNN speech recognition model.

The larger the Number of Filters of the CEL is, the better the word accuracy will be achieved for the LVCSR model. If you desire an effective Sound-2-Vector Embedding in your CEL+RNN architecture LVCSR model, then the Number of Filters of 100 or more is recommended for the Convolutional Embedding Layer, based on our experimental results in this study.

With the assistance from the Embedding Effect of the CEL, the training efficiency and the word accuracy of the models with CEL are significantly higher than that of the models without CEL.

The "Sound-2-Vector Embedding" End-to-End Speech Recognition System may require at least the following four key components:

- The Convolutional Embedding Layer (CEL) with appropriate parameters to optimize the feature extractions from sound signals with wide variations to embedded vectors.
- The CTC (Connectionist Temporal Classification) concept [1]: the direct target function used for optimizing the whole deep-learning neural network.
- Bidirectional GRU or LSTM in RNN: to process the sequence data.
- Dropout as an effective regulation.

The above developed Sound-2-Vector Embedding LVCSR system which has only 4.4 million model weight parameters, can be useful for mobile devices, Driverless Vehicles, and many other applications requiring **light-weighted high-accuracy** LVCSR models.